\documentclass[letterpaper, 10 pt, conference]{ieeeconf}
\IEEEoverridecommandlockouts
\let\labelindent\relax
\usepackage{enumitem}
\usepackage{balance}
\usepackage[font={small}]{caption}
\usepackage[labelformat=simple]{subfig}
\usepackage{array}
\usepackage{textcomp}
\usepackage{mathtools, nccmath}
\usepackage{graphicx}
\usepackage{amsfonts}
\usepackage{amsmath}
\usepackage{amssymb}
\usepackage{hyperref}
\usepackage{tikz}
\usepackage{arydshln}
\usepackage{multirow}
\usepackage{bm}
\usepackage{epstopdf}
\usepackage{cite}
\usepackage{siunitx}
\usepackage{xcolor}
\usepackage[ruled,linesnumbered,boxed,noend]{algorithm2e}
\usepackage{booktabs}
\usepackage{graphicx}
\usepackage{threeparttable}
\usepackage{stfloats}

\title{\LARGE \bf 
Range-SLAM: Ultra-Wideband-Based Smoke-Resistant Real-Time Localization and Mapping
}
\author{ Yi Liu$^{*}$, Zhuozhu Jian$^{*}$, Shengtao Zheng, Houde Liu$^{\dag}$, Xueqian Wang, Xinlei Chen$^{\dag}$, Bin Liang
\thanks{* contributed equally; $^\dag$ corresponding authors.}
\thanks{
This paper was supported by the Yunnan Forestry and Grassland Science and Technology Innovation Joint Special Project (grant NO. 202404CB090017), the Natural Science Foundation of China under Grant 62371269, 92248304, Shenzhen Science Fund for Distinguished
Young Scholars under Grant RCJC20210706091946001, Guangdong Innovative and Entrepreneurial Research Team Program (2021ZT09L197), and Meituan.}
\thanks{Yi Liu, Zhuozhu Jian, Shengtao Zheng, Houde Liu, Xueqian Wang, and Bin Liang are with Shenzhen International Graduate School, Tsinghua University, Shenzhen 518055, China, \tt\{yiliu24@mails., jzz24@mails., st\_zheng24@mails., liu.hd@sz., wang.xq@sz., liangbin@\}tsinghua.edu.cn}
\thanks{Xinlei Chen is with the Shenzhen International Graduate School, Tsinghua University, Shenzhen 518055, China, Pengcheng Lab, Shenzhen 518055, China, RISC-V International Open Source Laboratory, Shenzhen 518055, China e-mail: 
\href{mailto:chen.xinlei@sz.tsinghua.edu.cn}{chen.xinlei@sz.tsinghua.edu.cn}.
}
}
\begin{document}
\maketitle
\begin{abstract}
This paper presents Range-SLAM, a real-time, lightweight SLAM system designed to address the challenges of localization and mapping in environments with smoke and other harsh conditions using Ultra-Wideband (UWB) signals. While optical sensors like LiDAR and cameras struggle in low-visibility environments, UWB signals provide a robust alternative for real-time positioning. The proposed system uses general UWB devices to achieve accurate mapping and localization without relying on expensive LiDAR or other dedicated hardware. By utilizing only the distance and Received Signal Strength Indicator (RSSI) provided by UWB sensors in relation to anchors, we combine the motion of the tag-carrying agent with raycasting algorithm to construct a 2D occupancy grid map in real time. To enhance localization in challenging conditions, a Weighted Least Squares (WLS) method is employed. Extensive real-world experiments, including smoke-filled environments and simulated scenarios, demonstrate the efficiency and robustness of the proposed system. 
\end{abstract}

\section{Introduction}
\label{sec:Introduction}

LiDAR and camera are commonly used external sensors for agent positioning and mapping\cite{1,2,3}. However, in challenging scenes filled with smoke or containing large numbers of mirrors, the relatively low signal-to-noise ratio of these optical measurement devices allows SLAM algorithms that rely on such devices to malfunction, significantly reducing the capabilities of autonomous systems \cite{cao2021tare}\cite{jian2022putn}. Over the few years, there have been various studies to explore radio frequency (RF) bands instead of dedicated sensors for SLAM, given the RF signals' robustness against harsh and extreme environmental conditions. Moreover, compared to WiFi \cite{kim2024structure} and sonar \cite{santos2014fusing} signals, UWB signal offers significant advantages in terms of both accuracy and robustness \cite{14}. While novel RF devices such as UWB radar\cite{15} and micro-Doppler radar\cite{23} are capable of initially accomplishing SLAM tasks in challenging environments, their relatively high cost makes it difficult for them to be widely adopted in industrial and production experiments. Therefore, we aim to design a real-time, lightweight, onboard algorithm that can perform localization and mapping tasks solely based on general UWB devices.

Accomplishing SLAM tasks with general UWB devices presents two key challenges: 1) UWB sensors provide only distance and RSSI measurements from anchors, limiting the ability to construct a dense geometric map of the environment. 2) In non-line-of-sight (NLOS) conditions, UWB-based localization suffers from significant inaccuracies due to signal obstructions and reflections\cite{12}\cite{13}. 
To address these challenges, this paper first proposes a SLAM system that relies solely on UWB devices for real-time localization and mapping. First, we develop a low-complexity, real-time NLOS identification module based on Support Vector Machines (SVM) to detect obstacles between anchors and tags. Next, we combine the robot's motion data with a raycasting algorithm to construct a 2D occupancy grid map. Finally, we enhance localization accuracy in NLOS areas using WLS based on the constructed grid map.



\begin{figure}[t]
    \centering
    \includegraphics[width=8.5cm]{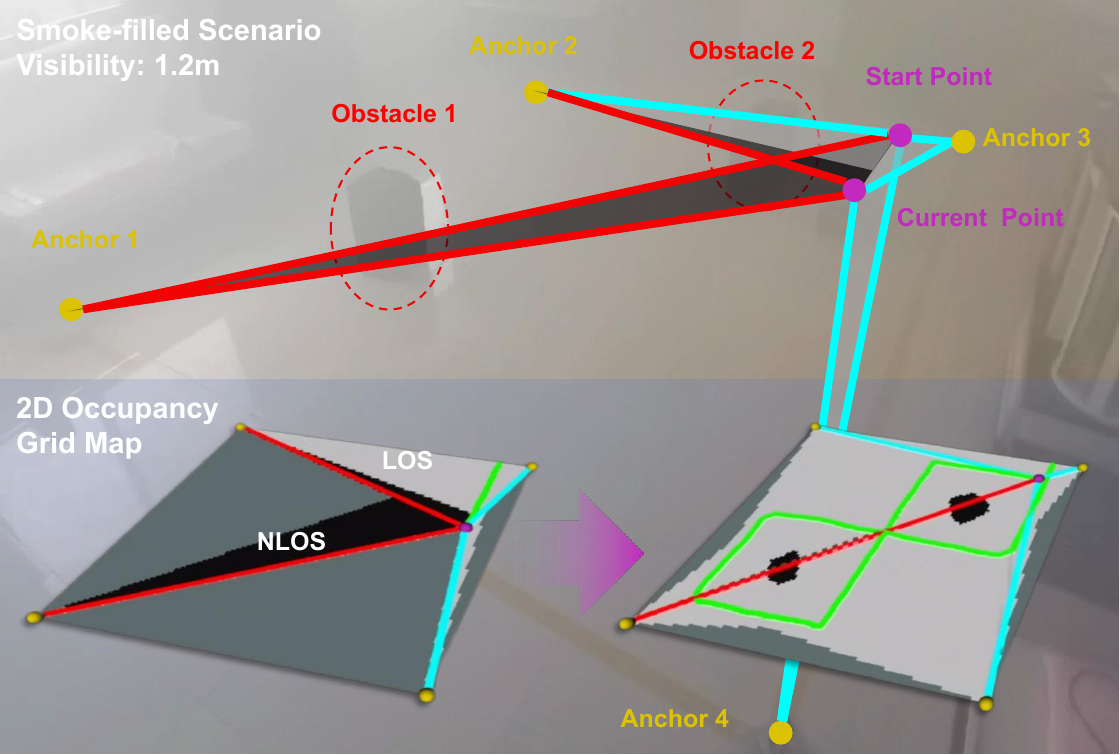}
    \caption{Range-SLAM enables robots to perform environment mapping and localization at 50Hz in smoke-filled environments.}
    \label{first}
    \vspace{-0.2cm}
\end{figure}

\begin{figure*}[t]
    \center
    \includegraphics[width=17.5cm]{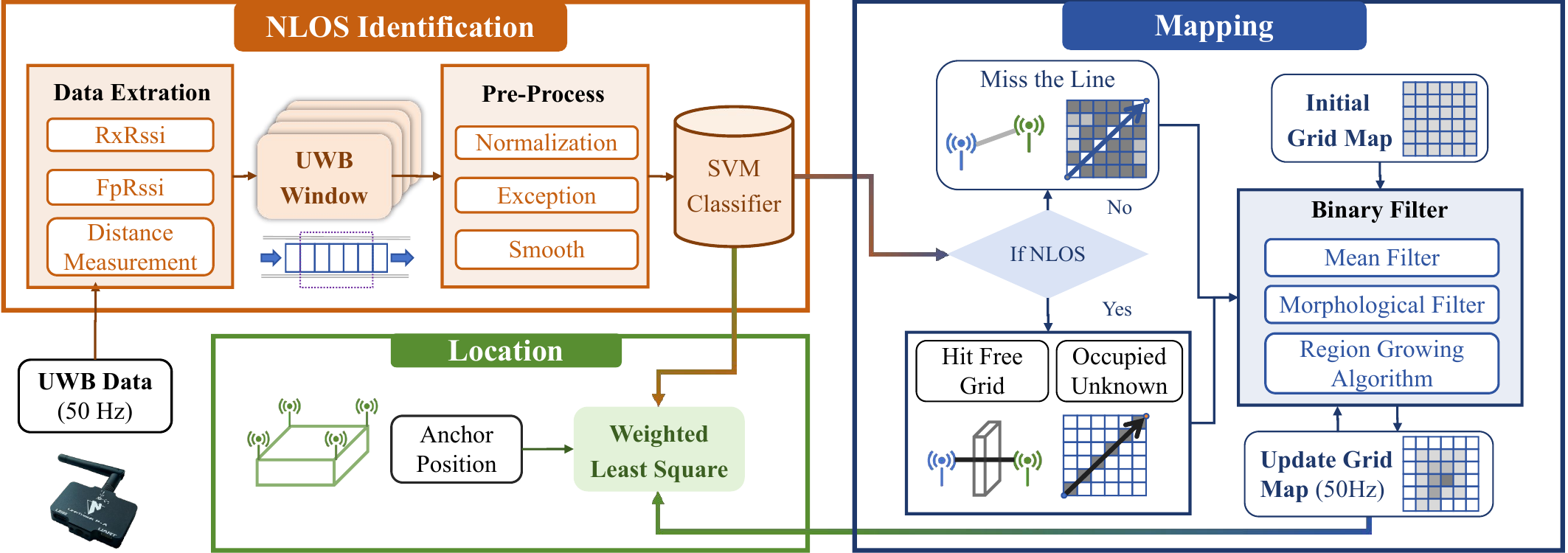}
    \caption{
    From left to right: the ``NLOS Identification" module \ref{subsec:NLOS Identification} performs UWB data (50Hz) extraction, including distance measurement, RxRssi, and FpRssi, followed by preprocessing steps such as normalization, exception handling, and smoothing. An SVM classifier is applied to identify the LOS condition. Subsequently, the ``Mapping" module \ref{subsec:Mapping Algorithm} utilizes a raycasting algorithm to continuously update the grid map. Finally, the ``Location" module \ref{subsec:Location Algorithm} use the WLS method to estimate the tag's position based on the grid map.}
    \label{Fig:framework}
    \vspace{-0.3cm}
\end{figure*}

\subsection{Related Work}
In environments filled with smoke, dust, or strong reflections, UWB signals have been shown to possess stronger anti-interference capability and higher transmission rates compared to LiDAR and RGB cameras \cite{ding2018research}. However, in these environments, challenges remain in achieving accurate localization and obstacle mapping using UWB devices.

Current research on UWB localization focuses on addressing the attenuation issues in NLOS scenarios. Multi-sensor fusion \cite{chen2022improved, xunt2023crepes, yang2021novel} and the use of redundant UWB base stations \cite{al2014improved}\cite{li2023comprehensive} can provide the system with more observations, improving localization accuracy. However, due to limitations in cost and portability, hardware conditions in real-world scenarios often fail to meet these requirements. 

Machine learning methods are often used for identifying NLOS and LOS conditions. Once these conditions are identified accurately, other techniques are used to correct errors caused by NLOS. Guvenc et al. \cite{guvenc2007nlos} introduced a WLS algorithm that works better than LS in NLOS situations using Monte Carlo and TOA data. Li et al. \cite{li2017three} proposed a UWB algorithm that improves accuracy by 20-30\% over LS. Dong et al.\cite{dong2023uwb} developed IRACKF to reduce NLOS errors with a special covariance matrix, boosting UWB accuracy. However, these methods struggle with low RSSI attenuation objects (e.g., Wooden Plate, Concrete Wall) or environments with smoke and dust, leading to reduced localization accuracy. Our method enhances identification capability by collecting historical UWB detection data through the online construction of a 2D probabilistic grid map, effectively improving accuracy in challenging environments. To accurately construct obstacle information in the environment, we propose a method that combines the motion of UWB tags with raycasting for map construction and updating.
\subsection{Contributions}
This work offers the following contributions: 
\begin{enumerate}
    \item We propose a time-efficient, lightweight mapping framework that integrates the motion of a tag-carrying agent with a raycasting algorithm, relying solely on UWB ranging information and RSSI to construct a 2D occupancy grid map in real time.
    \item We propose a method that combines the WLS algorithm with the constructed grid map, significantly improving the localization accuracy of UWB systems in scenarios where NLOS identification capability is diminished.
    \item Simulation and real-world experiments are conducted to validate the real-time capabilities, efficiency, and robustness of the mentioned algorithm.

\end{enumerate}

\section{Overview of the Framework}
\label{sec:framework}

\subsection{Problem Statement}
\label{subsec:problem}
\
In the indoor scenario, there are UWB anchors with known fixed locations denoted as $P_{A_i}\in \mathbb{R} ^2\left( i=1...N_A \right) $ while the mobile agent carries a UWB tag, where $N_A$ is the total number of UWB anchors. Denote the input data of each UWB anchor $\textit{i}$ at time $\textit{t}$ as $D_{i}^{t}:=\left[ d_{i}^{t}, Rx_{i}^{t}, Fp_{i}^{t} \right] ^T(i=1...N_A)$, including the distance measurement $d^t_i\in \mathbb{R}$, the residual received signal strength indicator $Rx^t_i\in \mathbb{R}$, and the first path received signal strength indicator $Fp^t_i\in \mathbb{R}$. 

Define $\mathbf{X}\subset\mathbb{R}^2$ as the work space. 
$\mathbf{X}_{obs}, \mathbf{X}_{free}\subset\mathbf{X}$ is denoted as obstacle and non-obstacle area of space respectively. Specifically, we can set up $M_{Map}\times N_{Map}$ occupancy grids map $\textbf{M}=\{m_{j,k}\}(j=1...M_{Map},\textit{k}=1...N_{Map})$ representing $\mathbf{X}$, where each grid contains a value $-1$(obstacle) or $1$(non-obstacle). Define $\textbf{p}^t(\textit{x,y})$ as the estimated 2D position of UWB tag at time $\textit{t}$.

The formal definition of the problem is as follows: Given observed UWB dataset $D^t$,$(i.e. \{D_1^t,D_2^t...D_N^t\})$, the task is to estimate the position $\textbf{p}^t(\textit{x,y})$ and update the state of $\textbf{M}$. The $\textbf{M}$ and $\textbf{p}^t(\textit{x,y})$ should satisfy: 1) representing the surrounding obstacle $\mathbf{X}_{obs}$. 2) minimizing root mean squared error for the position vector. 3) minimizing the time needed to complete the SLAM.\par

\subsection{System Framework}
\
The structure of the proposed Range-SLAM algorithm completes real-time indoor localization and mapping tasks through a series of interconnected processes, as shown in Fig.\ref{Fig:framework}. The system is mainly divided into three modules, which are: the ``NLOS Identification" module \ref{subsec:NLOS Identification}, the ``Mapping" module \ref{subsec:Mapping Algorithm} and the ``Location" module \ref{subsec:Location Algorithm}.



\section{Implementation}
\label{sec:implementation}

\subsection{NLOS Identification}
\label{subsec:NLOS Identification}

\subsubsection{Data Pre-Process}
\ 
\newline
\indent 
The data are collected from UWB signals by a mobile tag, including measured distance $d$, RxRssi $Rx$, and FpRssi $Fp$ to identify NLOS conditions, as briefly represented in $[*]$. To reduce noise, we apply a UWB window that retains the latest $N_W$ frames. The data is first processed through normalization:



\begin{equation}
    \label{equ:distribution}
	\begin{aligned}
\mathbf{\tilde{[*]}}_{i}^{t}=\left( \mathbf{[*]}_{i}^{t}-\mu_{[*]} \right) /\sigma_{[*]} 
	\end{aligned}
\end{equation}
where $\mu_{[*]}\in \mathbb{R}$ and $\sigma^2_{[*]} \in \mathbb{R}$ are the mean and the covariance of the trained UWB dataset $D^{train}$ respectively. Then, outliers are removed using an exception-based filtering process to eliminate points with excessive deviations.
\begin{equation}
    \label{equ:exception}
	\begin{aligned}
(\tilde{{[*]}}^t_i-\mu_{w,[*]})\in[-3\sigma_{w,[*]},-3\sigma_{w,[*]}]
\end{aligned}
\end{equation}
where $\mu_{w,[*]}\in \mathbb{R}$ and $\sigma^2_{w,[*]} \in \mathbb{R}$ respectively represent the mean and the covariance of the UWB window. Finally, we smooth the data:
\begin{subequations}\label{eqn-4}
  \begin{align}
    &\tilde{[*]}^t_{i,S}=\sum\nolimits_{i=1}^{N_S} k_n \tilde{[*]}^{t-n}_{i} \\
    s.t.  \quad &\sum\nolimits_{n=1}^{N_S}k_n= 1 \quad k_n > k_{n+1}
  \end{align}
\end{subequations}
where $k_i$  denotes the weight for the $i$-th lagged data point, $\tilde{[*]}^{t-n}_{i}$ is the data point value at time $t-n$ within the UWB window, $\tilde{[*]}^t_{i,S}$ stands for the weighted moving average of the UWB window.

\subsubsection{SVM Model}
\ 
\newline
\indent Identifying LOS/NLOS by signal strength and distance is a classification problem. This paper utilizes SVM classifier to address it, as detailed below:
\begin{subequations}\label{eqn-4}
  \begin{align}
    &\min\limits_{{\omega},{e}} \quad ||{\omega}||_2^2+C(\sum\nolimits_{i}\nolimits^{N} e_i)\\
    s.t. \quad &\forall i \ , \ y^{(i)}{\omega}^T{X^{(i)}}\leq 1-e_i \quad e_i > 0
  \end{align}
\end{subequations}
where $X^{(i)}=[d^{(i)},Fp^{(i)},Rx^{(i)},1]^T$ denotes the support vector, and $y^{(i)}$ is the manually annotated output label ($1$ for LOS, $-1$ for NLOS). The training vector $\omega=[\omega_1,\omega_2,\omega_3,b]^T$  learns from $N$ samples. The model is regularized by $C(\sum\nolimits_{i}\nolimits^{N} e_i)$, adjusting $C$ for better generalization and optimizing  deviations $e_i$ for ideal classification. Given that UWB data follows a Gaussian distribution, we apply the Gaussian kernel\cite{16} in our method. During prediction, a smoothed sample $\tilde{D}_{i,S}^{t}=\left[ d_{i,S}^{t},Rx_{i,S}^{t},Fp_{i,S}^{t} \right] ^T$ is converted to $x^{test}:=[{\tilde{D}_{i,S}^{tT}},1]^T$ and input into the trained SVM model:

\begin{equation}
    \label{equ:SVM}
	\begin{aligned}
y^{test}=w^T x^{test}
\end{aligned}
\end{equation}
where the result $y^{test}$ denotes the output score obtained from the test set through the SVM.\par 
\begin{figure*}[t]
    \centering
    \includegraphics[width=17.5cm]{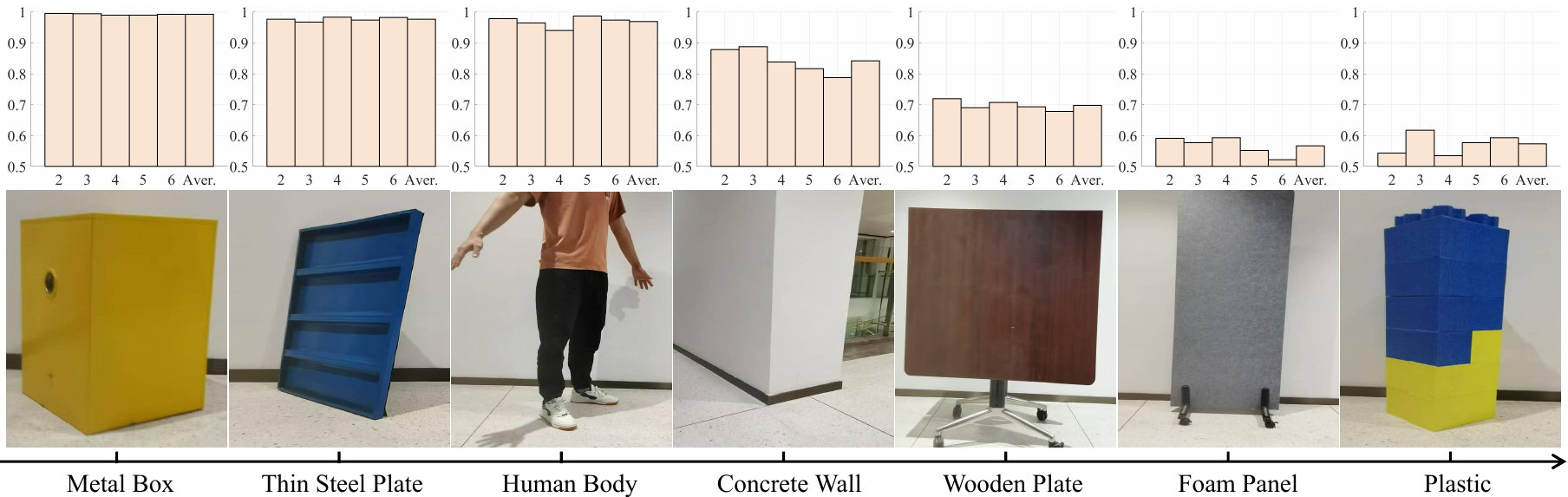}
    \caption{\textbf{The NLOS identification accuracy comparison within the different ranges for seven obstacles.} In each bar chart, the y-axis represents the recognition accuracy rate, while the x-axis denotes the distance between the tag and anchor. To address NLOS identification, we use an SVM classifier for real-time binary classification, achieving near-perfect NLOS recognition for iron materials and human bodies with significant positioning interference, some recognition rate is sacrificed for materials like plastic and foam yet.}
    \label{NLOS}
    \vspace{-0.2cm}
\end{figure*}
Prior to Range-SLAM experiments, we rigorously assessed the SVM binary classifier's NLOS recognition accuracy for seven obstacle types as Fig.\ref{NLOS}. Data collected at $1m$ intervals from $2-6m$ include dynamic movements and extreme obstacle edges for realism.

\SetKwFor{For}{for}{\string do}{}
\RestyleAlgo{ruled}
\SetKwFor{For}{for}{\string do}{}
\RestyleAlgo{ruled}
\begin{algorithm}[t]
    \caption{Range-SLAM($D^t=\{D_1^t...D_N^t\},\textbf{M}$)}
    \label{alg:PF-RRT*}
    \LinesNumbered
    
    $W.\text{initial}()$\;
    \For{\rm \textit{i=} 1 to \textit{N}}
    {
        $\tilde{D}^t_i = \textbf{Normorlization}(D^t_i) $\;
            $W_i.\text{update}(\hat{D}^t_i)$\;
            \uIf{$\tilde{D}^t_i \textless \textbf{Exception}(W)$}{$\KwRet$\;}
            $\hat{D}^t_i = W_i.\textbf{smooth}(\tilde{D}^t_i)$\;
            $[\text{LOS}_i,\text{score}_i ]= \textbf{SVM}.\text{input}(\hat{D}^t_i)$\;
            $\beta_i=\textbf{Uniform}(score_i)$\;
            $\textbf{e}_{NLOS,i}^t = \textbf{NLOS\_Matching}(\beta_i)$\;
    }
    $\textbf{e}_{Map}^t = \textbf{Prior\_Map\_Matching}(\textbf{M})$\;

    $\textbf{e}_{Mot}^t = \textbf{Motion\_Model}(\hat{\textbf{p}}^{t-1},\textbf{u}^{t-1})$\;

    $\textbf{p}^t = \textbf{Optimization}(\textbf{e}_{NLOS},\textbf{e}_{Mot}^t,\textbf{e}_{Map}^t)$\;
    \For{\rm \textit{i=} 1 to \textit{N}}
    {
        $\textbf{M}_{line}=\textbf{RayCasting}(\textbf{p}^t,\textbf{M})$\;
        \uIf{$LOS_i==1$}{$\textbf{M}_{pro}=\textbf{Free}(\textbf{M}_{line})$\;}
        \uElse
        {
            $\textbf{M}_{pro}=\textbf{occupy}(\textbf{M}_{unknown})$\;
        }
    }
    $\tilde{\textbf{M}}=\textbf{Sgn\_Function}(\textbf{M}_{pro})$\;
    $\textbf{M}=\textbf{Binary\_Filter}(\tilde{\textbf{M}})$\;
    \KwRet $\textbf{M},\textbf{p}^t$,
\end{algorithm}

\subsection{Mapping Algorithm}
\label{subsec:Mapping Algorithm}
\subsubsection{Raycasting}
\label{subsec:Grid}
\begin{figure}[t]
    \centering
    \includegraphics[width=8.5cm]{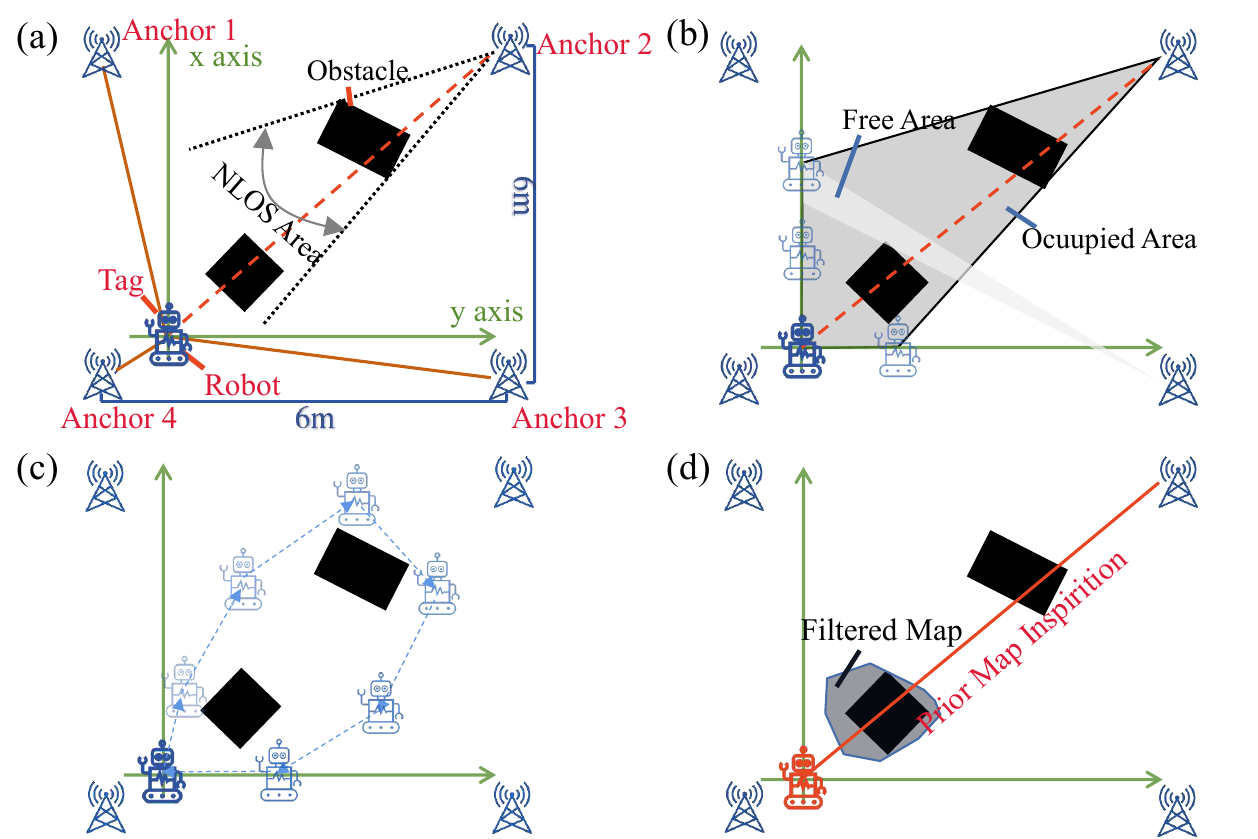}
    \caption{Illustration of Range-SLAM mapping and localization enhancement. (a)Scene Description (b)Free-Occupied Mapping Algorithm (c)Motion Real-time Mapping (d)UWB Prior Map for Localization Enhancement}
    \label{UWB}
    \vspace{-0.2cm}
\end{figure}

\
\newline
\indent
Since UWB only provides range information without other information, it is necessary to integrate robot mobile movement. Therefore, we introduce Raycasting for mapping to integrate grid data with LOS information in maps. As shown in Fig.\ref{UWB}(a) red dashed line, it starts from the tag's view, tracing to an anchor. Bresenham's algorithm\cite{17} accurately plots the line, identifying grid intersections. This determines the grids $\textbf{M}_{line}=\{m_{i,j}\}_{i=n...N}^{j=k...K} \subset \textbf{M}$ through which LOS or NLOS passes on the corresponding map.

\subsubsection{Free-Occupied Mapping}
\label{Free-Occupied Mapping}
\
\newline
\indent 
When a clear LOS path is established between a UWB anchor and tag, it means there are no obstructions, so we classify those grid as Free, as shown in Fig.\ref{UWB}(b) Free Area. Conversely, it is shown in Fig.\ref{UWB}(b) Occupied Area.\par
However, when observing an LOS/NLOS situation, we cannot definitively conclude that all grid cells along the path necessarily contain obstacles or non-obstacles due to the accuracy of nlos identification. To address this problem,we perform a ``hit grid” method to update the ``free” and ``occupy" grid as the following formula:\par
\begin{subequations}\label{eqn-4}
  \begin{align}
    &\{\textbf{M}_{line}\}_{LOS}\ \ +=P_{free}\\
    &\{\textbf{M}_{line}\}_{NLOS}-=P_{occupy}
  \end{align}
\end{subequations}
where $\{*\}_{LOS}$ represents the LOS area and $\{*\}_{NLOS}$ represents the NLOS area that has been scanned. $P_{free}$ and $P_{occupy}$ denotes the corresponding incremental value.\par
Following multiple rounds of collision hit verifications as shown in Fig.\ref{UWB}(c), we progressively update the status of the grid based on accumulating evidence. we ultimately process this probability map using the $sgn(x)$ function:
\begin{subequations}\label{eqn-4}
  \begin{align}
    &sgn(x)=\left\{
\begin{aligned}
&\quad 1          &x>0\\
&-1            &x<0   \\
\end{aligned}
\right.\\
    &\quad\quad \textbf{M}=sgn(\textbf{M}_{pro})
  \end{align}
\end{subequations}
where $\textbf{M}_{pro}$ represents the probability map generated through the Free-Occupied Mapping algorithm and $\textbf{M}$ is the final binary map. To refine the binary grid map, we implemented an enhanced binary filtering technique\cite{18}\cite{19}, which includes mean filtering, morphological filtering,
and a region-growing algorithm.

\subsection{Location Algorithm}
\label{subsec:Location Algorithm}
In order to obtain accurate location, we reformulate the UWB localization problem into a optimization as follows:
\begin{equation}
    \label{equ:motion model}
	\begin{aligned}
 \underset{\textbf{x}^t=(\textbf{p}^t,\textbf{u}^t)}{\operatorname{argmin}}&\left\{\rho_1  \left\|\textbf{e}_{NLOS}^t\right\|^2_2+\rho_2  \left\|\textbf{e}_{Mot}^t\right\|^2_2+\rho_3 \left\|\textbf{e}_{Map}^t\right\|^2_2\right\}
        \end{aligned}
\end{equation}
where $\boldsymbol{\rho}=\left[\rho_1,\rho_2,\rho_3\right]$ represents the weights of individual error terms $\textbf{e}_{NLOS}^t$, $\textbf{e}_{Mot}^t$, $\textbf{e}_{Map}^t$. $\textbf{p}^t=(x^t,y^t)$ is the estimated position and $\textbf{u}^t=(v_x^t,v_y^t)$ means the estimated input. Minimizing is accomplished via Newton-Gauss\cite{bin2015newton} or Levenberg-Marquardt\cite{ranganathan2004levenberg} method for speed. 
\subsubsection{NLOS Error Term}
\ 
\newline
\indent 
To take the NLOS identification results into account, the NLOS error term will be constructed as follows:
\begin{subequations}
  \begin{align}
    \textbf{e}_{NLOS,i}^t &= \beta_i \left(\tilde{d}^t_{i,S}-\left\|\textbf{p}^t-p_{A_i}\right\|\right)^2\\
    \quad\quad\beta _i &= \frac{1}{2}\left( 1+\tanh \left( \lambda y^{test} \right) \right) \label{beta}
  \end{align}
\end{subequations}
where $\beta_i$  is the weight for the ranging measurement of $i$ anchor, which is related to the score of SVM compution, $\lambda$ is a constant coefficient. 

\subsubsection{Motion Filter}
\ 
\newline
\indent 
To smooth the effects of discrete points obtained from UWB, we add a motion error term to flatten the results:
\begin{equation}
    \label{equ:weight2}
	\begin{aligned}
        \textbf{e}_{Mot}^t=
            \begin{bmatrix}
            \textbf{p}^t-f(\textbf{p}^{t-1},\textbf{u}^{t-1}) \\
            \textbf{u}^t-\textbf{u}^{t-1}
            \end{bmatrix}_{4 \times 1} 
    \end{aligned}
\end{equation}
We set the motion model as a uniform velocity model, specifically represented as follows:
\begin{equation}
    \label{equ:motion model}
	\begin{aligned}
f(\textbf{p}^{t-1},\textbf{u}^{t-1})=\textbf{p}^{t-1}+\textbf{u}^{t-1}\cdot \Delta t+\omega_t
\end{aligned}
\end{equation}
where $\omega_t$ represents the Gaussian noise during the motion process and $\Delta t$ denotes the time interval.

\subsubsection{Prior-Map Error Term}
\ 
\newline
\indent 
As shown in Fig.\ref{UWB}(d) red solid line,  when the robot passes through the mapped scene again, it is inspired by the prior map independently constructed by UWB:
\begin{subequations}
  \begin{align}
    \textbf{e}_{Map,i}^t &= \alpha_i \left(\tilde{d}^t_{i,S}-\left\|\textbf{p}^t-p_{A_i}\right\|\right)^2\\
    \quad\quad\alpha _i &=  1-\zeta \frac{N_{Occupy,i}}{N_{Total,i}}
  \end{align}
\end{subequations}
where $N_{Total,i}$ and $N_{Occupy,i}$ respectively represent the number of total grids and obstacle grids on the connection line between anchor $i$ and label, and $\zeta\in[0,1]$ represents the degree to which positioning is affected by the map.

\section{EXPERIMENTS}
To validate the performance of the algorithm, both simulation environments and real-world experimental platforms were constructed. In low-visibility environments, such as those filled with smoke, it is challenging to obtain ground truth, making it difficult to quantify localization and mapping accuracy. Therefore, we first conducted real-world experiments in a smoke-free environment to assess the impact of different numbers of agents and base stations on the algorithm. Next, we simulated complex scenarios in the virtual environment to test the robustness of our method under conditions where NLOS identification capability is reduced. Finally, experiments in smoke-filled environments were conducted to visually demonstrate the algorithm's performance in such conditions.
\subsection{Real-world Experiments}
\label{subsec: Real-world Experiments}

\subsubsection{Experiment Setup}

\begin{figure}[t]
    \centering
    \includegraphics[width=8cm]{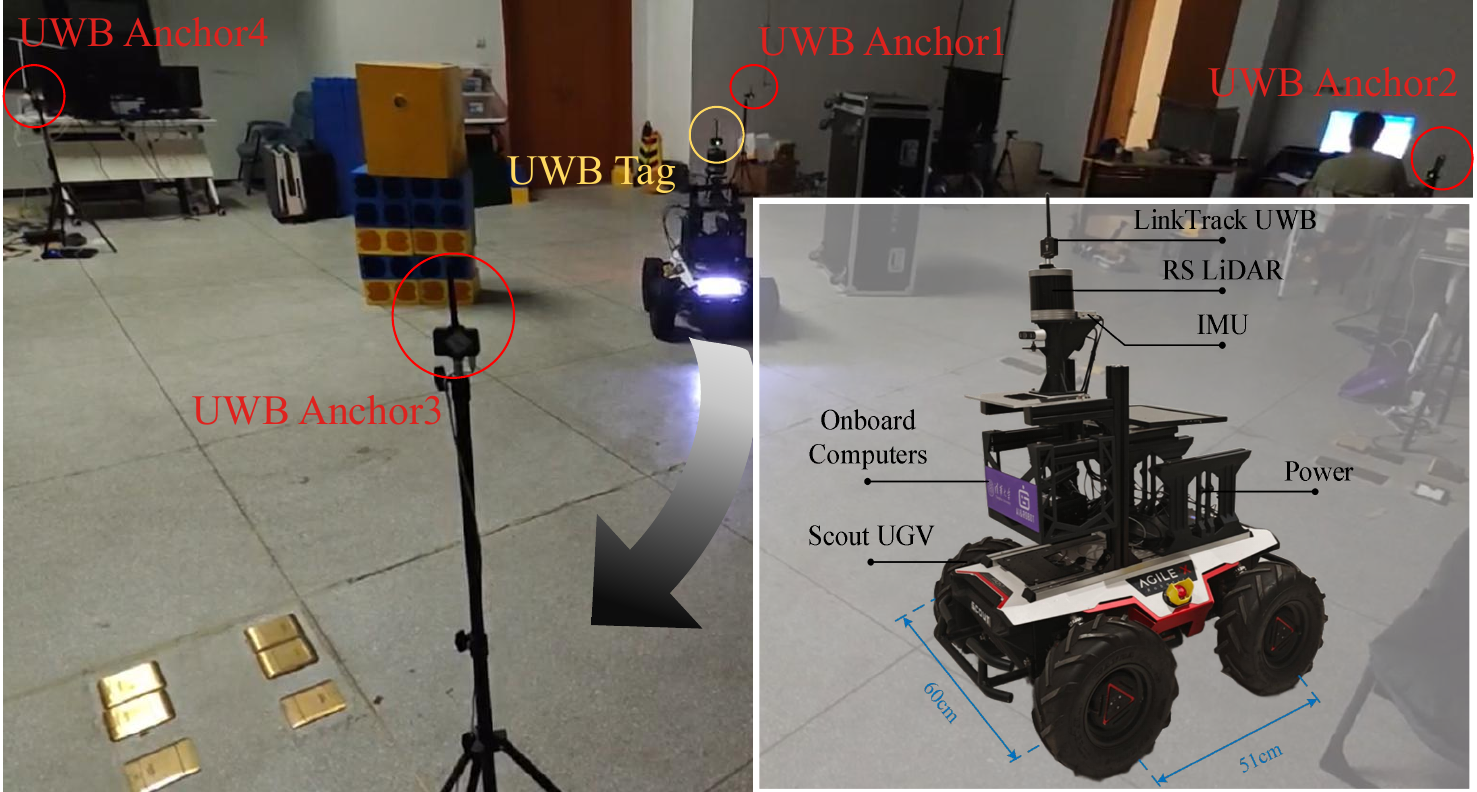}
    \caption{Our Range-SLAM mobile platform and real scenarios for the experiment.}
    \label{robot}
    \vspace{-0.2cm}
\end{figure}

As shown in Fig. \ref{robot}, we use the Scout2.0, a four-wheel-drive mobile robot, for the experiment. The UWB device is the NoopLoop LinkTrack, the IMU sensor is the WHEELTEC N200 IMU, the LiDAR sensor is the RS-Helios 32, and the computational hardware consists of an Intel NUC with an i5 2.4 GHz CPU and 16 GB of memory. The experimental area is set as a $6 m$ by $6 m$ square, with NoopLoop base stations deployed at each corner. The operator manually controls the robot to navigate through the area, following different paths around obstacles. 

\begin{figure}[t]
    \centering
    \includegraphics[width=8cm]{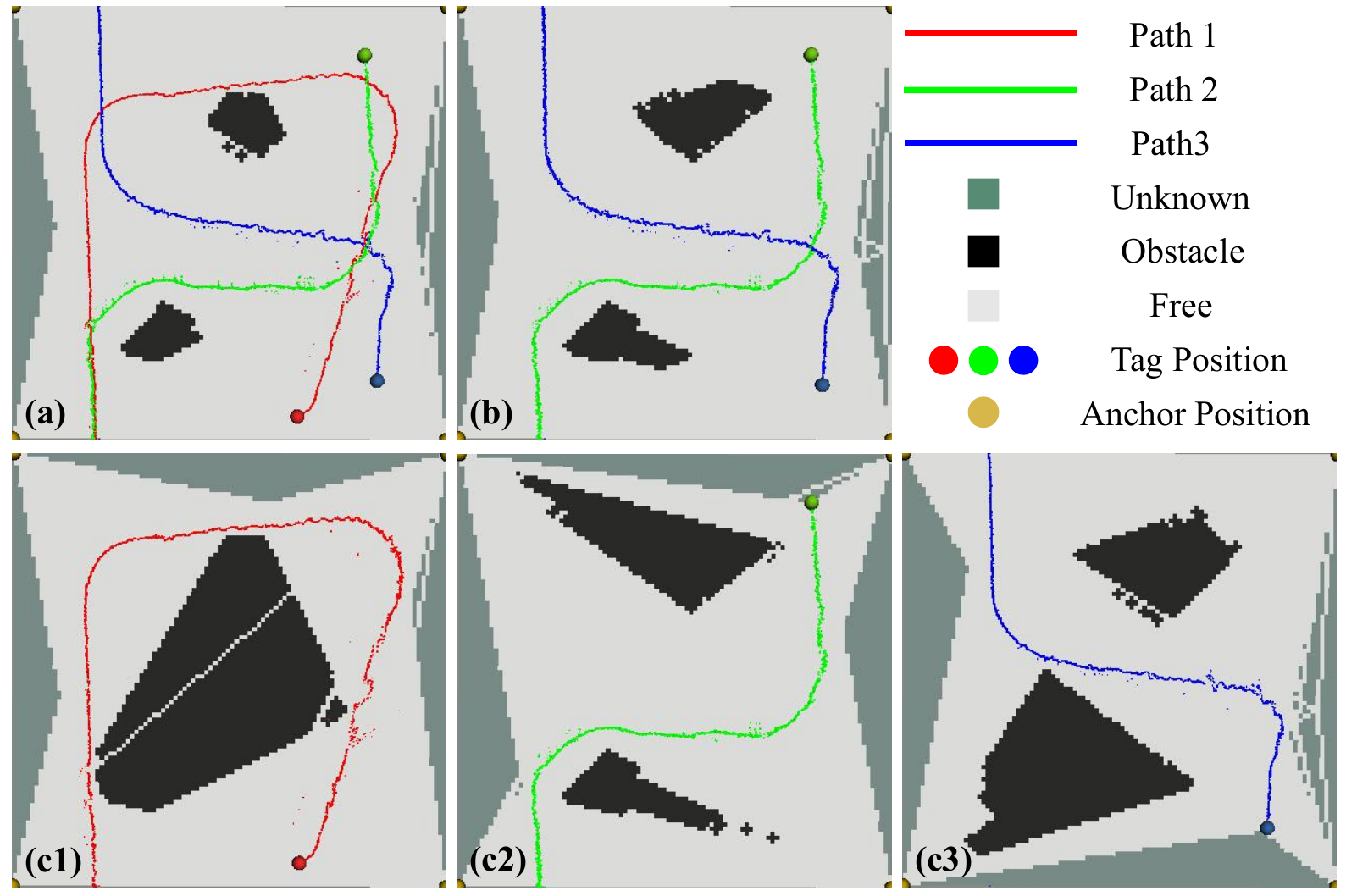}
    \caption{\textbf{Real-world Experiments for Multi Agents.} (a) and (b) respectively illustrate the performance of the multi-agent system and the dual-agent system, while the three graphs in (c) depict the performance of three distinct paths.}
    \label{multi}
    \vspace{-0.2cm}
\end{figure}

\begin{table}[h]  
  \centering  
  \begin{tabular}{ccccc}  
    \toprule
    System & Accuracy & Recall & F1 & ATE RMSE(cm) \\  
    \midrule
    Multi-System & 0.99 & 0.98 & 0.82  & 8.57 \\  
    Dual-System & 0.93 & 0.99 & 0.45 &9.31\\  
    Path-1 & 0.78 & 1  & 0.20 & 7.99\\  
    Path-2 & 0.83 & 1  & 0.26 & 8.03\\  
    Path-3 & 0.80 & 1  & 0.23 & 9.71\\  
    \bottomrule
  \end{tabular}
  \caption{\textbf{Range-SLAM for Different Motion Paths.}
  }  
  \label{multi_mapping}  
\end{table}

\subsubsection{Evaluation}
\
\newline
\indent
We test both single-robot and multi-robot configurations to evaluate the impact on map construction performance. And LiDAR and IMU-based FAST-LIO2 \cite{xu2022fast} id used as the ground truth. We use following metrics for evaluation\cite{accuray}:

\begin{table}[h]  
  \centering  
  \label{table5} 
    \begin{tabular}{|c|c|}   
    \hline
    \textbf{Indicators} & \textbf{Meaning} \\
    \hline
      TP & Obstacles correctly explored \\  
    \hline
      TN & Non-obstacles correctly explored\\  
    \hline
      FP & Obstacles falsely explored\\  
    \hline
      FN & Non-obstacles falsely explored \\  
    \hline
     Accuracy & Correctly identifying both obstacles and non-obstacles\\
    \hline
        Recall & Ability to detect actual obstacles among all obstacles\\
    \hline
     F1 Score & Balance in identifying obstacle and non-obstacle
    \\
    \hline
        RMSE & The root mean square of absolute positioning errors\\
        \hline
    \end{tabular}
    \caption{\textbf{Mapping and Localization Metrics.}}
\end{table}

\begin{figure}[!t]
    \centering
    \includegraphics[width=8cm]{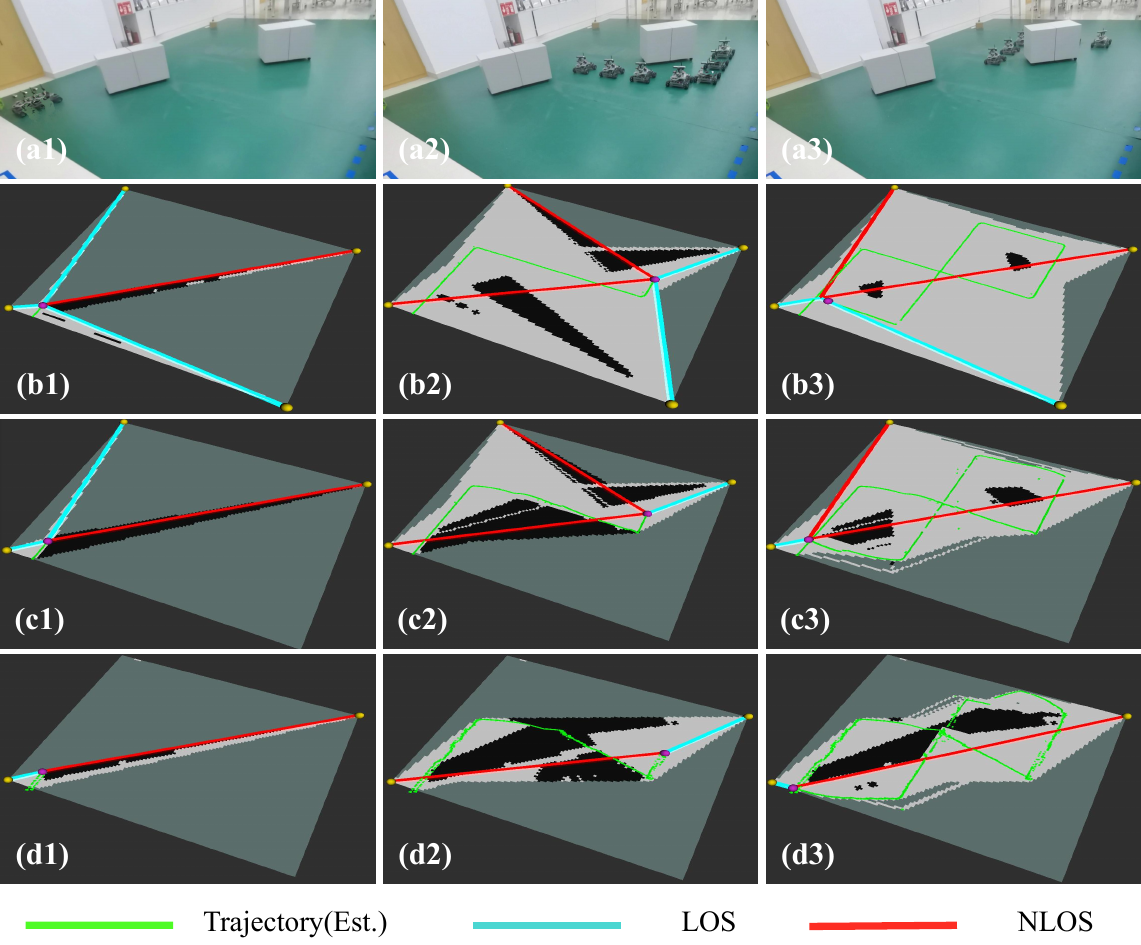}
    \caption{\textbf{Ablation study with different anchor num.} In the experiments in (b), (c) and (d), 4, 3, and 2 anchors are used, respectively.}
    \label{different_achor}
    \vspace{-0.2cm}
\end{figure}

As shown in Fig. \ref{multi} and TABLE \ref{multi_mapping}, for the single-robot, different motion paths result in varying geometric structures of the constructed obstacles, while the mapping performance of multiple agents is significantly better than single agent. The variation in the number of base stations has an even more significant impact on mapping performance. In Fig.\ref{different_achor},  we run the algorithm with different anchor number. As the number of anchors decreases, trajectory deviations increase, and mapping performance significantly declines.


During algorithm execution, the UWB signal reception frequency is $50Hz$. Processing a UWB frame takes $0.529ms$ for localization, $0.206ms$ for NLOS identification, and $0.181ms$ for mapping. The CPU usage is $3.26\%$, and memory consumption is $8.8MB$. The experiments demonstrate that our method has minimal computational resource requirements and provides high real-time performance.

\begin{figure*}[t]
    \center
    \includegraphics[width=17.5cm]{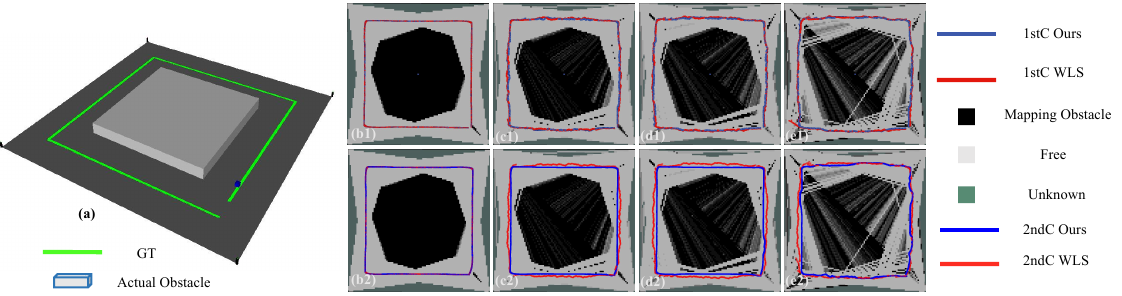}
    \caption{\textbf{Localization accuracy with NLOS identification capability.} The figure on the left (a) shows the simulated scenario. The figures from (b) to (e) on the right illustrate the performance with NLOS identification rates of 99\%, 80\%, 70\%, and 60\%, respectively. The red line represents the baseline, and the blue line shows the performance of our Range-SLAM approach. The upper part(1) shows the results for the first lap, while the lower part(2) presents the results for the second lap. }
    \label{sim}
    \vspace{-0.3cm}
\end{figure*}


\subsection{Simulation}
Different obstacle materials and smoke-filled environments can lead to reduced NLOS identification capability. 
We establish a GAZEBO simulation, as shown in Fig.\ref{sim}(a), with $4$ UWB stations in a $20\times20m$ area, centered around a $10\times10m$ obstacle. A tagged agent moves in two circular paths at a speed of $2.5m/s$ around a $15\times15m$ rectangle, sampled at $50Hz$. We extract the tag's position from GAZEBO as the ground truth. For comparison, we design a WLS algorithm in which the weights depend solely on real-time NLOS identification, without utilizing the 2D grid map.

\begin{figure}[t]
    \centering
    \includegraphics[width=8.5cm]{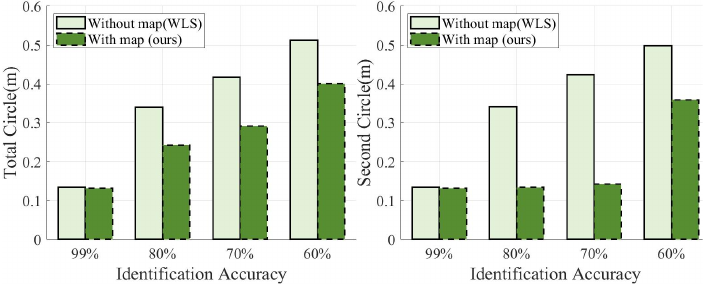}
    \caption{Comparison of our Range SLAM and WLS baselines in the overall and second circle ATE RMSE(m) indicators.} 
    \label{sim_localization}
    \vspace{-0.2cm}
\end{figure}

As shown in Fig.\ref{sim} and TABLE\ref{tab:metrics_table},  as the front-end NLOS identification capability decreases, both localization and mapping performance decline. As shown in Fig.\ref{sim_localization}, with the same front-end NLOS identification, Range-SLAM shows no significant improvement in localization accuracy over WLS in the first lap. However, in the second lap, Range-SLAM's accuracy improves notably, as the map from the first lap provides prior information, enhancing precision.

\begin{table}[t]
  \centering
  \begin{tabular}{ccccccccc}
    \toprule
    Percent & TN   & TP   & FN   & FP    & Accuracy & Recall & F1    \\ 
    \midrule
    99\%       & 6312 & 2500 & 0 & 1188       & 0.88   & 1      & 0.81  \\
    80\%       & 6296 & 2500 & 0 & 1204       & 0.87   & 1      & 0.80 \\
    70\%       & 6160 & 2477 & 23 & 1340   & 0.86   & 0.99 & 0.78 \\
    60\%       & 5560 & 1928 & 572 & 1940   & 0.76   & 0.77 & 0.61 \\
    \bottomrule
  \end{tabular}
  \caption{\textbf{Performance metrics across different percentages.} The table presents the TP, TN, FP, FN, Precision, Accuracy, Recall, and F1 Score for each percentage level, showing how performance metrics vary as the percentage decreases.}
  \label{tab:metrics_table}
  \vspace{-0.2cm}
\end{table}

\begin{figure}[t]
    \centering
    \includegraphics[width=7cm]{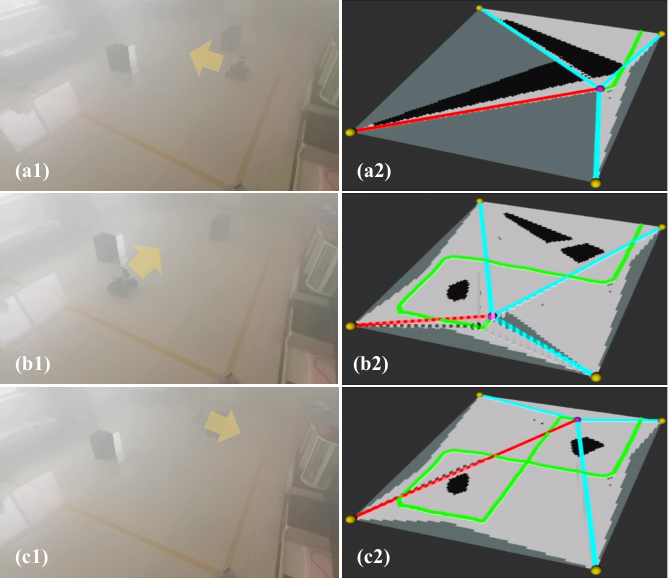}
    \caption{Mapping and Localization Performance in a Smoky Environment.} 
    \label{smoke}
    \vspace{-0.2cm}
\end{figure}

\subsection{Smoke-filled Scenario}
We run Range-SLAM in a smoke-filled environment with a visibility of only 1.2 meters. The robot is controlled to follow a figure-eight trajectory. As the robot moves, Range-SLAM demonstrates strong mapping and localization capabilities despite the challenging conditions in Fig.\ref{smoke}.

\section{Conclusion and Future Work}
\label{sec:conclusion}
This study proposes Range-SLAM, a real-time and lightweight method using RSSI and Range Information to achieve mapping and enhance positioning with UWB sensors solely. A noval positioning method based on prior maps constructed by UWB independently has been proposed, designed to cope with extreme smoke scenarios where the accuracy of NLOS identification significantly deteriorates. However, this mapping methods rely on operator-controlled robot movement. In the future, we aim to develop autonomous navigation algorithms to actively construct a complete environmental map.



\newpage
{
    \balance
    \bibliographystyle{IEEEtran}
    \bibliography{IEEEabrv, bib/bibliography}
}

\end{document}